\documentclass[11pt,a4paper]{article}
\usepackage[hyperref]{emnlp-ijcnlp-2019}
\usepackage{times}
\usepackage{latexsym}

\usepackage{amsmath}
\usepackage{amssymb}
\usepackage{ctable}
\usepackage{enumitem}
\usepackage{pifont}
\usepackage{makecell}
\usepackage{url}
\usepackage{graphicx}
\usepackage{footnote}
\usepackage{tablefootnote}

\makesavenoteenv{tabular}
\makesavenoteenv{table}
\graphicspath{{images/}}

\aclfinalcopy % Uncomment this line for the final submission

\newcommand{\xmark}{\text{\ding{55}}}

\newcommand{\Ours}{CoSQL}
\newcommand{\syncon}{SyntaxSQL-con}

\newcommand{\seqcon}{CD-Seq2Seq}
\newcommand*{\affaddr}[1]{#1} % No op here. Customize it for different styles.
\newcommand*{\affmark}[1][*]{\textsuperscript{#1}}
\newcommand*{\email}[1]{\texttt{#1}}

\newcommand{\vic}[1]{\textbf{\textcolor{magenta}{Vic: #1}}}

\newcommand{\hide}[1]{}

\title{\Ours{}: A Conversational Text-to-SQL Challenge Towards Cross-Domain Natural Language Interfaces to Databases}

\author{
Tao Yu\affmark[$\dagger$] 
\quad Rui Zhang\affmark[$\dagger$]
\quad He Yang Er\affmark[$\dagger$] 
\quad Suyi Li\affmark[$\dagger$] 
\quad Eric Xue\affmark[$\dagger$] 
\quad Bo Pang\affmark[$\dagger$]
\\{\bf Xi Victoria Lin\affmark[$\P$]
 \quad Yi Chern Tan\affmark[$\dagger$]
 \quad Tianze Shi\affmark[$\S$]
\quad Zihan Li\affmark[$\ddagger$] 
\quad Youxuan Jiang\affmark[$\ddagger$]} 
\\{\bf Michihiro Yasunaga\affmark[$\dagger$]
\quad Sungrok Shim\affmark[$\dagger$]
\quad Tao Chen\affmark[$\dagger$]
\quad Alexander Fabbri\affmark[$\dagger$]} 
\\{\bf Zifan Li\affmark[$\dagger$]
\quad Luyao Chen\affmark[$\ddagger$]
\quad Yuwen Zhang\affmark[$\ddagger$]
\quad Shreya Dixit\affmark[$\dagger$]
\quad Vincent Zhang\affmark[$\dagger$]}
\\{\bf Caiming Xiong\affmark[$\P$]
\quad Richard Socher\affmark[$\P$]
\quad  Walter S. Lasecki\affmark[$\ddagger$]
\quad Dragomir Radev\affmark[$\dagger$]}
\\
\affaddr{\affmark[$\dagger$]Yale University} \affaddr{\affmark[$\P$]Salesforce Research}\\
\affaddr{\affmark[$\ddagger$]University of Michigan} \affaddr{\affmark[$\S$]Cornell University}\\
\email{\{tao.yu,\,r.zhang,\,dragomir.radev\}@yale.edu}\\
\email{\{xilin,\,cxiong,\,rsocher\}@salesforce.com}
}

\date{}

\begin{document}
\maketitle
\begin{abstract}
We present \Ours{}, a corpus for building cross-domain, general-purpose database (DB) querying dialogue systems.
It consists of 30k+ turns plus 10k+ annotated SQL queries, obtained from a Wizard-of-Oz (WOZ) collection of 3k dialogues querying 200 complex DBs spanning 138 domains.
Each dialogue simulates a real-world DB query scenario with a crowd worker as a user exploring the DB and a SQL expert retrieving answers with SQL, clarifying ambiguous questions, or otherwise informing of unanswerable questions. 
When user questions are answerable by SQL, the expert describes the SQL and execution results to the user, hence maintaining a natural interaction flow.
\Ours{} introduces new challenges compared to existing task-oriented dialogue datasets: (1) the dialogue states are grounded in SQL, a domain-independent executable representation, instead of % predefined 
domain-specific slot-value pairs, and (2) because testing is done on unseen databases, success requires generalizing to new domains. %due to testing on unseen databases.
\Ours{} includes three tasks: SQL-grounded dialogue state tracking, response generation from % SQL and result
query results, and user dialogue act prediction.
We evaluate a set of strong baselines for each task and show that \Ours{} presents significant challenges for future research.
The dataset, baselines, and leaderboard will be released at \url{https://yale-lily.github.io/cosql}.

\end{abstract}

\section{Introduction}

\begin{figure}[!t]
    \vspace{-1.5mm}\hspace{-1mm}
    \centering
    \includegraphics[width=0.48\textwidth]{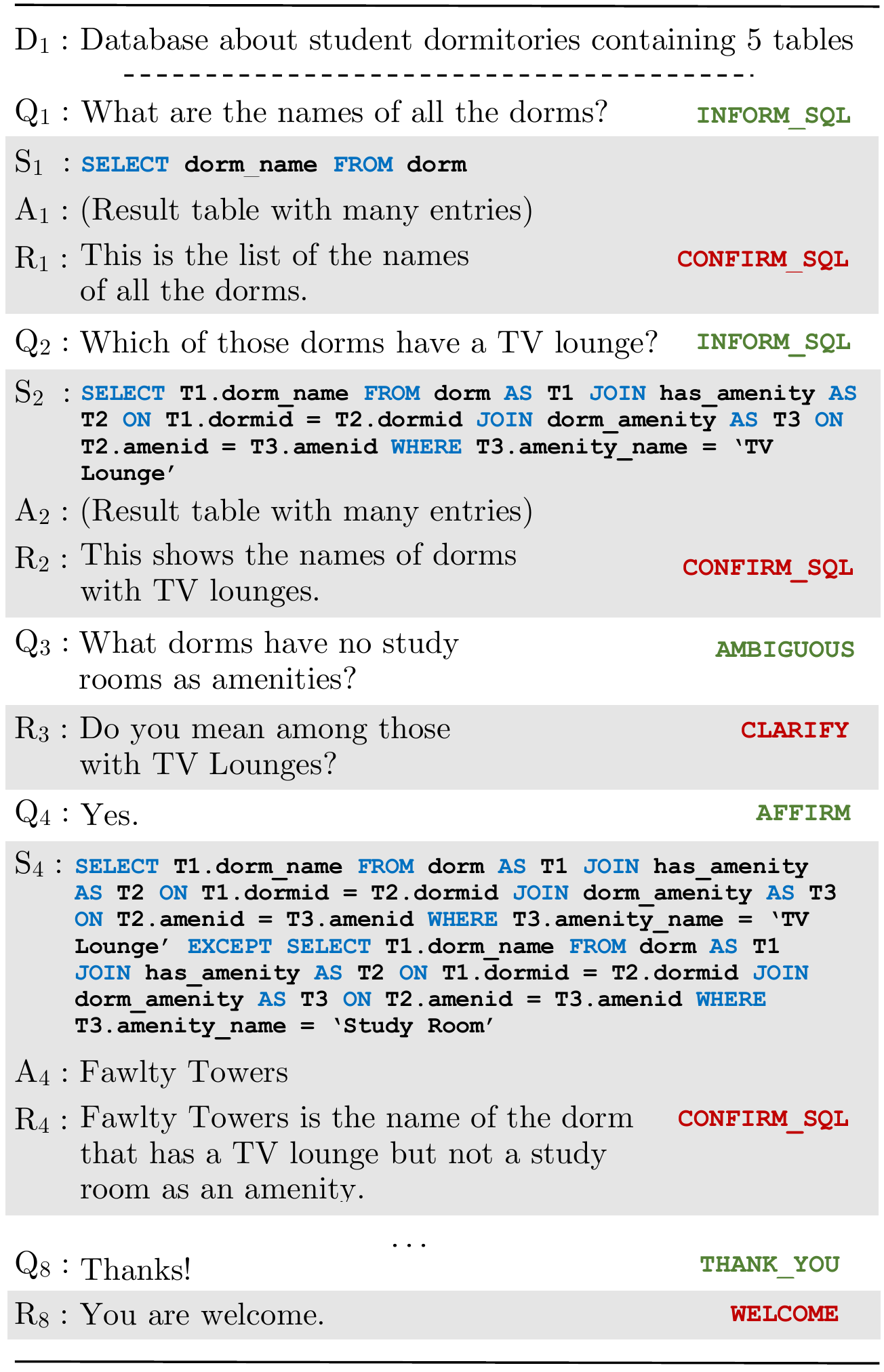}\vspace{-2mm}
    \caption{
    A dialog from the \Ours{} dataset. Gray boxes separate the user inputs ($Q_i$) querying the database ($D_i$) from the SQL queries ($S_i$), returned answers ($A_i$), and expert responses ($R_i$). Users send an input to the expert, who writes the corresponding SQL query (only seen by the expert) if possible and sends an answer and response description back. Dialogue acts are on the right-hand side (e.g., $Q_3$ is ``ambiguous" and $R_3$ is ``clarify").
    }
\label{fig:task}
\vspace{-6mm}
\end{figure}

%\hide{
Natural language interfaces to databases (NLIDB) have been studied extensively, with a multitude of different approaches introduced over the past few decades.
To this end, considerable progress has been made in querying data via natural language (NL).
However, most NL query systems expect the query to be well-formed and stated in a single sentence \cite{zelle96,li2014constructing,yaghmazadeh2017sqlizer,iyer17,Zhong2017,Xu2017,shi2018incsql,2018executionguided,Yu&al.18.emnlp.syntax,Yu&al.18.emnlp.corpus}.
In reality, complex questions are usually answered through interactive exchanges (Figure~\ref{fig:task}).
Even for simple queries, people tend to explore the database by asking multiple basic, interrelated questions \cite{Hale06,Levy08,Frank13,Iyyer17}.
This requires systems capable of sequentially processing conversational requests to access information in relational databases.
To drive the progress of building a context-dependent NL query system, corpora such as ATIS \cite{Hemphill90,Dahl94} and SParC \cite{Yu2019}\footnote{SParC task is available at \url{https://yale-lily.github.io/sparc}} have been released.
However, these corpora assume all user questions can be mapped into SQL queries and do not include system responses.

Furthermore, in many cases, multi-turn interaction between users and NL systems is needed to clarify ambiguous questions (e.g., $Q_3$ and $R_3$ in Figure~\ref{fig:task}), verify returned results, and notify users of unanswerable or unrelated questions.
Therefore, a robust dialogue-based NL query agent that can engage with users by forming its own responses has become an increasingly necessary component for the query process.
Such systems have already been studied under task-oriented dialogue settings by virtue of continuous effort of corpus creation \cite{Seneff2000,Walker2002,Raux2005LetsGP,Mrksic2015MultidomainDS,AsriSSZHFMS17,Budzianowski2018MultiWOZA} and modelling innovation \cite{Artzi2011,henderson2013,lee2016,Su2016OnlineAR,Dhingra2016EndtoEndRL,Li2016DeepRL,MrksicSWTY16}.
The goal of these systems is to help users accomplish a specific task, such as flight or hotel booking or transportation planning.
However, to achieve these goals, task-oriented dialogue systems rely on pre-defined slots and values for request processing (which can be represented using simple SQL queries consisting of \texttt{SELECT} and \texttt{WHERE} clauses).
Thus, these systems only operate on a small number of domains and % as a result of their task-specific slot-value pairs ,
have difficulty capturing the diverse semantics of practical user questions.% likely to be encountered in practice.

In contrast, the goal of dialogue-based NLIDB systems is to support general-purpose exploration and querying of databases by end users.
To do so, these systems must possess the % following capabilities: the 
ability to (1) detect questions answerable by SQL, (2) ground user questions into executable SQL queries if possible, (3) return results to the user in a way that is easily understood and verifiable, and (4) handle unanswerable questions.
The difficulty of constructing dialogue-based NLIDB systems stems from these requirements.
%Developing such systems introduces several challenging tasks, including conversational query specification, access to multiple target information sources, and comprehensible and reliable result presentation.
To enable modeling advances in this field, we introduce \Ours{}, the first large-scale cross-domain \textbf{Co}nversational text-to-\textbf{SQL} corpus collected under the WOZ setting~\cite{Budzianowski2018MultiWOZA}.
\Ours{} contains 3,007 dialogues (more than 30k turns with annotated dialogue acts and 10k expert-labeled SQL queries) querying 200 complex DBs spanning across 138 different domains.
For each dialogue, we follow the WOZ set-up that involves a crowd worker as a DB user and a college computer science student who is familiar with SQL as an expert (\S\ref{sec:data_collection}).

Like Spider\footnote{Spider task is available at \url{https://yale-lily.github.io/spider}} \cite{Yu&al.18.emnlp.corpus} and SParC \cite{Yu2019}, the % large number of domains 
cross-domain setting in \Ours{} enables us to test the ability of systems to generalize on querying different domains via dialogues.
We split the dataset in a way that each database only appears in one of train, development, or test set.
This setting requires systems to generalize to new domains without additional annotation.

More importantly, unlike most prior work in text-to-SQL systems, \Ours{} demonstrates greater language diversity and more frequent user focus changes.
It also includes a significant amount of questions that require user clarification and questions that cannot be mapped to SQL queries, introducing the potential to evaluate text-to-SQL dialog act prediction. These features pose new challenges for text-to-SQL systems.
%Question-SQL pairs in this corpus reflect greater diversity in user backgrounds compared to other corpora and involve frequent changes in user intent between pairs or ambiguous questions that require user clarification. These features pose new challenges for text-to-SQL systems.
Moreover, 
% as comprehensible and reliable presentation of query results is a crucial component of any NLIDB system, 
\Ours{} includes system responses that describe SQL queries and the returned results in a way that is easy for users with different backgrounds to understand and verify, as faithful and comprehensible presentation of query results is a crucial component of any NLIDB system.\footnote{The DB community has developed query visualization~\cite{li2014constructing} and other techniques to provide faithful explanation of a SQL query. These explanations are complementary to the NL ones used in our work and future NLIDB systems could integrate them.}

We introduce three challenge tasks on \Ours{}: (1) \textbf{SQL-grounded dialogue state tracking} to map user utterances into SQL queries if possible given the interaction history (\S\ref{sec:dstc}),
(2) \textbf{natural language response generation} based on an executed SQL and its results for user verification (\S\ref{sec:datg}) and (3) \textbf{user dialogue act prediction} to detect and resolve ambiguous and unanswerable questions  (\S\ref{sec:ddtg}).
% To show the challenges presented by \Ours{}, 
We provide detailed data analysis and qualitative examples (\S\ref{sec:data_stats_analysis}).
For each of the three tasks, we benchmark several competitive baseline models (\S\ref{sec:result}).
The performances of these models indicate plenty of room for improvement.

\section{Related Work}
\label{sec:rel}

% Some related work includes text-to-SQL generation \cite{Suhr:18context, Yu&al.18.emnlp.corpus}, task-oriented dialog systems \cite{Henderson2014}, and data-to-text generation \cite{Wiseman17}. (to double check references)

\paragraph{Text-to-SQL generation} 
Text-to-SQL generation %is an important subtask of semantic parsing that 
has been studied for decades in both DB and NLP communities. \cite{warren1982efficient,Zettlemoyer05,popescu2003towards,li2006constructing,li2014constructing,iyer17,Zhong2017,Xu2017,Yu18,dong18,cathy18,Guo2019TowardsCT,Bogin2019RepresentingSS}.
However, the majority of previous work focus on converting a single, complex question into its corresponding SQL query.
Only a few datasets have been constructed for the purpose of mapping context-dependent questions to structured queries.
\citet{Price90,Dahl94} collected ATIS that includes series of questions from users interacting with a flight database.
%While some works \cite{Miller96,Zettlemoyer09,Suhr:18context,zhang19} have used this dataset, ATIS only has a single domain for flight planning, and this limits the possible SQL clauses available.
\citet{Yu2019} introduced SParC, a large cross-domain semantic parsing in context dataset, consisting of 4k question sequences with 12k questions annotated with SQL queries over 200 complex DBs. Similar to ATIS, SParC includes sequences of questions instead of conversational interactions. An NL question and its corresponding SQL annotation in SParC are constructed by the same expert.
Recent works \cite{Suhr:18context,zhang19} have built context-dependent text-to-SQL systems on top of these datasets.
% However, there is no realistic dialogue between the user and the system involved.
% All questions and SQL queries are annotated by the same student with SQL experience based on Spider \cite{Yu&al.18.emnlp.corpus}.
% Other similar works such as DialSQL \cite{gur2018} and \citet{iyer17} ask a few fixed multiple choice questions to gain more information regarding what the correct SQL query should be and have their systems adjust accordingly.
% Therefore, there is no realistic dialogue between the user and the system involved.

In contrast, \Ours{} was collected under a WOZ setting involving interactions between two parties, which contributes to its diverse semantics and discourse covering most types of conversational DB querying interactions (e.g. the system will ask for clarification of ambiguous questions, or inform the user of unanswerable and irrelevant questions).
% Instead of assuming that each user question can be answered by a SQL query as in all other text-to-SQL datasets, dialogues in \Ours{} contain ambiguous, unanswerable, and irrelevant questionxdrrs from the user, in addition to more diverse intent switches.
Also, \Ours{} includes a natural language system response for the user to understand and verify the system’s actions.
% If the user’s question can be expressed as a SQL query, the system will utilize the SQL query and the returned result table to formulate a response describing the two. Otherwise, the system will ask for clarification in the case of ambiguous questions, or inform the user of unanswerable and irrelevant questions.

% \paragraph{Sequential Question Answering} To address the task of sequential question answering, \citet{Iyyer17} introduced the SequentialQA corpus, which was created by decomposing the complicated questions in WikiTableQuestions \cite{pasupat2015compositional} into sequences of simple, interrelated questions.
% However, as a result, SequentialQA only contains simple tables that are independent of one another.
% Furthermore, because the answers to the questions in SequentialQA must appear in the associated tables, most of them can be solved by simple \texttt{WHERE} conditions.

\paragraph{Task-oriented dialog systems} 
Task-oriented dialog systems \cite{Henderson2014,Wen16,mrksic17,Budzianowski2018MultiWOZA} have attracted increasing attention especially due to their commercial values.
The goal is to help users accomplish a specific task % that only spans at most a few domains, 
such as hotel reservation, flight booking, or travel information.
% To accomplish such goal, 
These systems \cite{BordesW16,zhong2018global,WuTradeDST2019} often pre-define slot templates grounded to domain-specific ontology, limiting the ability to generalize to unseen domains. 
In comparison, our work is to build a system for general-purpose DB exploration and querying. The domain-independent intent representation (SQL query) enables the trained system to work on unseen domains (DB schemas). % without defining new anthologies.
% It does not adopt prxdrredefined slot templates and does not actively poke the user for information except when it needs to clarify an ambiguous question.
% Most importantly, instead of predicting user goals and requests that are comprised of a set of domain-specific slot-value pairs, as in the dialog state tracking task of task-oriented systems, the dialogue states in \Ours{} are grounded to SQL.
%(also take system response as an input when user questions are ambiguous, In our task, it is possible that the possible context input spans more than one previous user utterances.)
% Furthermore, the division of the dataset based on DB ensures that no DB appears more than once among the training, development, and testing sets in \Ours{}, forcing systems that tackle our tasks to generalize to arbitrary domains.

While % the system responses in 
most task-oriented dialog systems need to actively poke the user for information to fill in pre-defined slot-value pairs, % and accomplish the specified task, 
the primary goal of system responses in \Ours{} is to offer users a reliable way to understand and verify the returned results.
If a question can be converted into a SQL query, the user is shown the execution result and the system will describe the SQL query % in the context of 
and the result in natural language.
In case s% the cases where 
the user questions are ambiguous or unanswerable by SQL, the system either requests the user to rephrase or informs them to ask other questions.
% Because of the differences usage, we define some new user and system dialog acts.
%The traditional dialogue state \cite{Henderson2014,mrksic17,zhong2018global} is presented as predefined slot-value pairs with limited natural language meaning representations. 
%\cite{emily2018}

\paragraph{Data-to-Text generation} 
Response generation in \Ours{} takes a structured SQL query and its corresponding result table to generate an NL description of the system's interpretation of the user request. % what the system understands the user's request to be. % for any arbitrary domain. 
Compared to most dialogue-act-to-text generation tasks, the richer semantics of SQL queries makes our task more challenging -- besides generating natural and coherent descriptions, faithfully preserving the logic of a SQL query in an NL response is also crucial in our task.
Furthermore, this component is related to previous work on text generation from structured data~\cite{McKeown1985,iyer-etal-2016-summarizing,Wiseman17}.

\section{Data Collection}
\label{sec:data_collection}

We follow the Wizard-of-Oz setup %, which involves facilitating 
which facilitates dialogues between DB users and SQL experts to create \Ours{}.
We recruited Amazon Mechanical Turkers (AMT) to act as DB users and trained 25 graduate- and undergraduate-level computer science students proficient in SQL to act as DB experts.
The collection interface \cite{lasecki2013conversations} is designed to be easy-to-operate for the experts and intuitive for the users.
Detailed explanations of the data collection process % from both sides 
is provided below.

\hide{
We follow the WOZ setup, which involves facilitating dialogues between crowd workers and computer science students, to create \Ours{}.
For each dialog, we employ two annotators, a DB user and an DB expert.
To capture the diversity of user backgrounds, we use Amazon Mechanical Turk (AMT) to recruit DB users.
However, because DB experts should have professional SQL backgrounds and consistent performances while answering the questions of DB users, AMT workers would not suffice to fill these roles.
Therefore, we trained 25 graduate- and undergraduate-level computer science students in SQL so that they can act as DB experts.
Unlike SParC, where questions in the same interaction are written and answered by the same student annotator, \Ours{} has an AMT worker (DB user) ask questions and a student (DB expert) answer them using SQL.
This allows us to capture a variety of behavior from DB users and collect natural dialog interactions.
The collection setup is designed to provide an easy-to-operate system interface for the Wizards and easy-to-follow goals for the users.
}

\paragraph{Reference goal selection}
\label{ref_goal_select}
We pre-select a reference goal for each dialogue to ensure the interaction is meaningful and to reduce redundancy within the dataset. 
Users are asked to explore the given DB content to come up with questions that are likely to naturally arise in real-life scenarios and reflect their query intentions as specified by the reference goals.
Following~\citet{Yu2019}, we selected the complex questions classified as medium, hard, and extra hard in Spider~\cite{Yu&al.18.emnlp.corpus} as the reference goals.\footnote{\citet{Yu2019} also includes 12.9\% of the easy questions in Spider in order to increase dataset diversity. In this work we prioritize the complex questions that trigger more interesting interactions and do not include any easy questions.}
% a cross-domain condext-independent text-to-SQL dataset annotated by 11 SQL experts, and use them as the reference goals. 
% \footnote{Users are allowed to ask other questions regarding the database and are highly encouraged to ask interrelated questions built upon what was previously asked.}
%\vic{Do we also include 12.9\% of the easy examples as SParC did? If so, we need to add the statement; otherwise, we need to modify 4,437 to the correct number.}
In total, 3,783 questions were selected on 200 databases.
After annotation and reviewing, 3,007 of them were finished and kept in the final dataset.

\paragraph{User setup}
\label{user_side}
We developed online chatting interfaces to pair the user with the expert (Figure \ref{fig:db_user_interface} and \ref{fig:db_user_related_questions} in Appendix).
When a data collection session starts, the user is first shown multiple tables from a DB to which a reference goal is grounded% related to the reference goal 
and is required to read through them.
Once they have examined the data stored in the tables, the reference goal question will be revealed on the same screen.
% They can either break down the given question into smaller questions or ask other related questions 
The user is encouraged to use the goal question as a guide to ask interrelated questions, % that refer to each other, 
but is also allowed to ask other questions exploring the DB.
We require the user to ask at least 3 questions.\footnote{The worker is paid \$1.20 USD for each dialog. To encourage interaction, we offer \$0.50 USD bonus for each dialogue if the user asks more than 4 interesting, interrelated questions.}
%, one of which should include the answer to the goal question.
In each turn, if the user question can be answered by a SQL query, they will be shown the result table, and the expert will write an NL response interpreting the executed SQL query based on their understanding of the user's query intent (Figure~\ref{fig:sql_expert_interface} Appendix).
If the user question is ambiguous or cannot be answered with SQL, they will receive clarification questions or notice to rephrase from the expert (detailed in expert setup).

\paragraph{Expert setup}
\label{system_side}
Within each session, the expert is shown the same DB content and the reference goal as the user (Figure \ref{fig:sql_expert_interface} in Appendix). 
For each dialogue turn, the expert first checks the user question and labels it using %as one of 
a set of pre-defined user dialog action types (DATs, see Table \ref{tb:acts}).
% \vic{add a figure/table defining the dialog action types and make a reference}. 
Then the expert sets the DAT of his response according to the user DAT. Both the user and the expert can have multiple DATs labels in each turn.
If the user question is answerable in SQL (labeled as \texttt{INFORM\_SQL}, e.g. $Q_1$ in Figure~\ref{fig:task} ), the expert writes down the SQL query\footnote{We use the same SQL annotation protocol as Spider \cite{Yu&al.18.emnlp.corpus} to ensure the same SQL pattern was chosen when multiple equivalent queries were available.}, executes it, checks the result table, and sends the result table to the user.
The expert then describes the SQL query and result table in natural language and sends the response.
If the user question is ambiguous, the expert needs to write an appropriate response to clarify the ambiguity (labeled as \texttt{AMBIGUOUS}, e.g. $Q_3$ in Figure~\ref{fig:task}).
Some user questions require the expert to infer the answer based on their world knowledge % require conducting more human inference based on their world knowledge and a query returned result
(labeled as \texttt{INFER\_SQL}, e.g. $Q_3$ in Figure~\ref{fig:dialogue_example_1}). 
If the user question cannot be answered by SQL, the expert will inform them to ask well-formed questions (labeled as \texttt{NOT\_RELATED}, \texttt{CANNOT\_UNDERSTAND}, or \texttt{CANNOT\_ANSWER}).
In other cases (labeled as \texttt{GREETING}, \texttt{THANK\_YOU}, etc.), the expert responds with general dialogue expressions ($Q_8$ in Figure~\ref{fig:task}).

\paragraph{User quality control}
Because of the real-time dialogue setting and the expensive annotation procedures on the expert side, conducting quality control on user is crucial for our data collection. 
We use LegionTools\footnote{\url{https://www.cromalab.net/LegionTools/}} \cite{Lasecki2014} to post our tasks onto AMT and to recruit and route AMT workers for synchronous real time crowd sourcing tasks. We specify that only workers from the U.S. with 95\% approval rates are allowed to accept our task.
Before proceeding to the chat room, each AMT worker has to go through a tutorial and pass two short questions\footnote{One is on how to ask interrelated questions and the other is on how to read multiple tables with reference keys.} to test their knowledge about our task.
% At the end of the tutorial, the worker is asked to take a survey of his background in data analysis, data tools, SQL, and databases.
Only the user who passes the quiz proceeds to the chat room.
%\footnote{We require each work to pass this process only once.}
Throughout the data collection, if a user ignores our instructions in a specific turn, we allow the experts to alert the user through chat and label the corresponding turn as \textsc{drop}.
If a user's actions continue to deviate from instructions, the expert can terminate the dialog before it ends.
After each dialogue session terminates, we ask the expert to provide a score from 1 to 5 as an evaluation of the user's performance.
Dialogues with a score below 3 are dropped and the user will be blocked from future participation.

\paragraph{Data review and post-process}
We conduct a multi-pass data reviewing process.\footnote{The review interface is shown in Figure \ref{fig:review_interface} (Appendix).}
Two student conducted a first-round review.
They focus on correcting any errors in the DATs of the users and the experts, checking if the SQL queries match the user's questions, and modifying or rewriting the expert's responses to contain necessary information in the SQL queries in case they miss any of them.
Also, they re-evaluate all dialogues based on the diversity of user questions and reject any dialogues that only contain repeated, simple, and thematically-independent user questions (about 6\% of the dialogs).
After the first-round review, another two student experts reviewed the refined data to double check the correctness of the DATs, the SQL queries, and the expert responses.
They also corrected any grammar errors, and rephrased the user's questions and the expert's responses in a more natural way if necessary.
Finally, we ran and parsed all annotated SQL queries to make sure they were executable, following the same annotation protocol as the Spider dataset.

\hide{
Each example was reviewed at least twice by three different groups of the students. 
We asked two students to conduct a first-round review.
More specifically, they focus on correcting any errors in the user's and the expert's dialog action types, checking if the SQL queries match the user's questions, and modifying or rewriting the expert's responses to contain all the information in the SQL queries.
Also, they re-evaluate all dialogues based on the diversity of user's questions and reject any dialogues that only contain repeated, simple, and thematically-independent user questions (estimate how many).
After the first-round review, another group of two students reviewed the reviewed data to double check the correctness of the dialog action types, the SQL queries, and the expert responses.
Moreover, they also corrected any grammar errors and rephrased the user's questions and the expert's responses in a more natural way if necessary (estimate how many).
Finally, we ran and parsed all annotated SQL queries to make sure they were executable and followed the Spider annotation protocol (estimate how many).
The review interface is shown in the Appendix.
}
\section{Data Statistics and Analysis}
\label{sec:data_stats_analysis}

We report the statistics of \Ours{} and compare it to other task-oriented dialog and context-dependent text-to-SQL datasets. We also conduct detailed analyses on its contextual, % dependencies
cross-domain nature, and question diversity.% , and provide some qualitative observations.
\hide{
In this section, we compare \Ours{} to other task-oriented dialogue and context-dependent text-to-SQL datasets.
Most importantly, the goal of our task is to help provide users with a general purpose tool that allows them query and explore arbitrary databases of their choosing rather than conduct domain specific tasks such as booking hotel rooms or taxis.
Compared to the predefined task-specific ontology and slot-value semantic representations found in other task-oriented dialog datasets, \Ours{} utilizes more general task settings and possesses powerful semantic representations based on cross-domain SQL queries of 200 different domains. 
%Because of this, the number of turns in our task is less than domain specific task oriented dialogue.
Moreover, compared to other context-dependent semantic parsing datasets such as SParC and ATIS, because \Ours{} does not assume that every user's question can be answered by SQL, it spans a greater diversity of question types, such as ambiguous questions and questions that cannot be directly answered or answered at all by SQL.
The sample from which the dialogues were collected incorporate users from a multiplicity of different backgrounds in the real world, further contributing to the corpus's diverse coverage.
Also, \Ours{} includes not only user questions but also assistant responses, which yields a more realistic and interactive task setting.
}

\begin{table*}[ht!]
\centering
\scalebox{0.8}{
\begin{tabular}{cccccc}
\Xhline{2\arrayrulewidth}
 & DSTC2 & WOZ 2.0 & KVRET & MultiWOZ & \textbf{\Ours{}} \\\hline
\# dialogs & 1,612 & 600 & 2,425 & 8,438 & 2,164\\
Total \# turns & 23,354 & 4,472 & 12,732 & 115,424 & 22,422\\
Total \# tokens & 199,431 & 50,264 & 102.077 & 1,520,970 & 22.8197\\
Avg. \# turns/dialog & 14.49 & 7.45 & 5.25 & 13.68 & 10.36 \\
Avg. \# tokens/turn & 8.54 & 11.24 & 8.02 & 13.18 & 11.34\\
Total \# unique tokens & 986 & 2,142 & 2,842  & 24,071  & 7,502\\
\# databases  & 1 & 1 & 1 & 7 & 140\\
\# Slots \# & 8 & 4 & 13 & 25 & 3,696\\
\# Values \# & 212 & 99 & 1,363 & 4,510 & $>$1,000,000\\
\Xhline{2\arrayrulewidth}
\end{tabular}}
\caption{Comparison of \Ours{} to some commonly used task-oriented dialogue datasets. The numbers are computed for the training part of data in consistency with previous work~\cite{Budzianowski2018MultiWOZA}.}
\label{tb:data_stats}
\end{table*}

\begin{table}[ht!]
\centering
\scalebox{0.78}{
\begin{tabular}{cccc}
\Xhline{2\arrayrulewidth}
   & \textbf{\Ours{}} & SParC & ATIS \\\hline
\# Q sequence  & 3,007 & 4298 & 1658 \\
\# user questions & 15,598$^*$ & 12,726 & 11,653 \\
\# databases & 200 & 200 & 1 \\
\# tables & 1020 & 1020 & 27 \\
Avg. Q len & 11.2 & 8.1 & 10.2 \\
Vocab & 9,585 & 3794 & 1582 \\
Avg. \# Q turns  & 5.2 & 3.0 & 7.0 \\
Unanswerable Q & \checkmark & \xmark & \xmark \\
User intent & \checkmark & \xmark & \xmark \\
System response & \checkmark & \xmark & \xmark \\
\Xhline{2\arrayrulewidth}
\end{tabular}}
\caption{Comparison of \Ours{} with other context-dependent text-to-SQL datasets. The number are computed over the entire datasets. $^*$For CoSQL we count the total \# user utterances.} 
\label{tb:data_stats_sp}
\end{table}

\paragraph{Data statistics}
\label{sec:data_stats}

\begin{figure}[!t]
    \centering
    \includegraphics[width=0.5\textwidth]{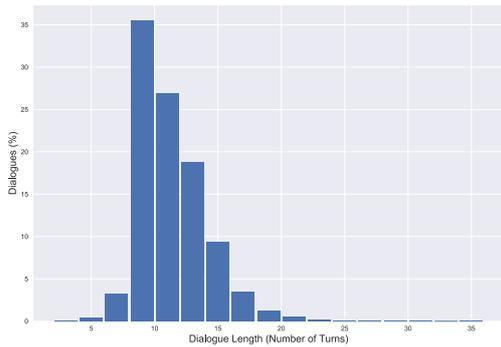}
    \vspace{-2mm}
    \caption{Distributions of dialogue lengths.
    %\vic{The axis marks of Figure 2,3,4 are too small.}
    }
\label{fig:data_stats}
\vspace{-2mm}
\end{figure}

Table \ref{tb:data_stats}å and \ref{tb:data_stats_sp} summarize the statistics of \Ours{}. 
\Ours{} contains 3k+ dialogues in total (2,164 in training), which is comparable to or bigger than most commonly used task-oriented dialogue datasets.
Figure \ref{fig:data_stats} shows the distribution of dialogue length in the corpus, approximately 80\% of dialogues involve 8 or more turns, with % an average of 11.21 turns per dialogue and 
a total of 31,148 turns.\footnote{Following~\citet{Budzianowski2018MultiWOZA}, in the statistics report we define the \# turns in a dialogue to be the total \# messages in the dialogue.}
The average number of tokens in each turn is 11.21. %, comparable to those of other task-oriented dialogue datasets.
Noticeably, the domain of CoSQL spans over 200 complex databases, overshadowing most other task-oriented dialogue datasets.
Comparing to existing context-dependent text-to-SQL datasets, \Ours{} contains significantly more turns, out of which 11,039 user utterances are convertible to SQL. In contrast, all NL utterances in ATIS and SParC can be mapped to SQL. \Ours{} also has a much larger NL vocabulary.
\hide{
Even though \Ours{} is not directly comparable to others because of the big difference in task goals, more than 3k dialogues were collected, as shown in Table \ref{tb:data_stats}, putting us ahead of most other task-oriented dialogue datasets in terms of size plus with much more possible slots and values \footnote{If we consider column names as slots and all DB entries as values.}
Comparing to prior context-dependent text-to-SQL datasets in Table \ref{tb:data_stats_sp}, the much larger numbers of \Ours{} in almost all dimensions demonstrate its diversity and difficulty.
}

\paragraph{Dialogue act distribution}
\label{sec:data_stats}

\begin{figure}[!t]
    \centering
    \hspace{-12mm}
    \includegraphics[width=0.5\textwidth]{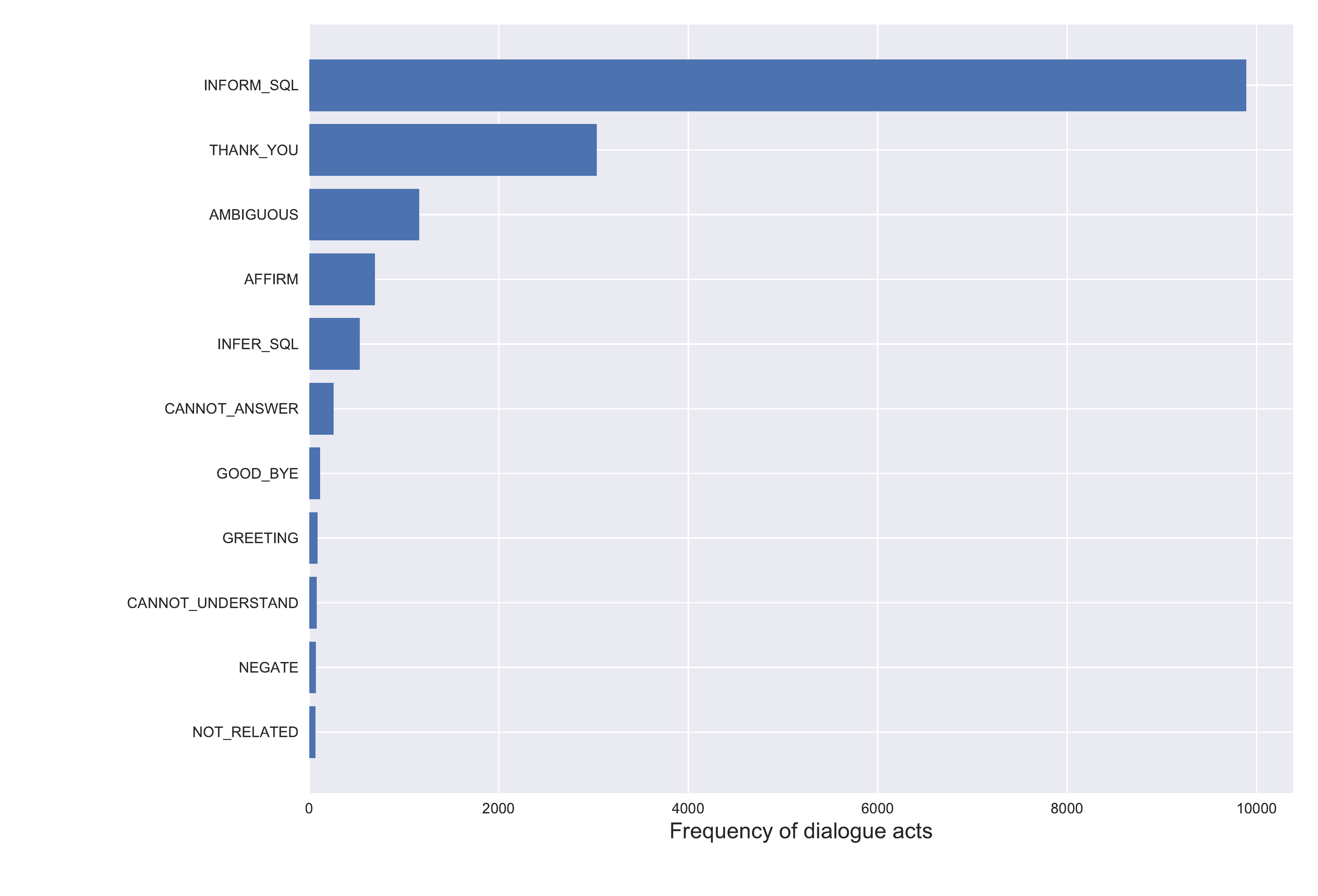}
    \vspace{-2mm}
    \caption{Distributions of user dialog action types.}% and percentages of dialogues containing special user dialog acts}
\label{fig:usr_labels}
\vspace{-2mm}
\end{figure}

As shown in Figure~\ref{fig:usr_labels}, \Ours{} contains a fairly diverse set of user dialogue action types (DATs). 
Unsurprisingly, \texttt{INFORM\_SQL} and \texttt{THANK\_YOU} are the two most commonly seen DATs. Among the rest of DATs, approximately 40\% are \texttt{AMBIGUOUS}, demonstrating the paramount importance of system clarification in % dialog-based semantic parsing and 
the DB querying process in general.
Another 20\% of this subgroup is \texttt{INFER\_SQL}, which signifies questions that cannot be answered without the aid of human inference.
% These proportions of \texttt{AMBIGUOUS} and \texttt{INFER\_SQL} identify them the most frequently occurring user dialogue acts outside of \texttt{INFORM\_SQL} and \texttt{THANK\_YOU}, testifying to the diverse complexity of \Ours{}.

\hide{
Because there are no restrictions on the questions the user can ask, dialogues in \Ours{} include those that cannot be answered by SQL using the available data, setting the corpus apart from other text-to-SQL datasets such as SParC and ATIS.
Of the user dialogue acts that are not \texttt{INFORM\_SQL} or \texttt{THANK\_YOU}, approximately 40\% are \texttt{AMBIGUOUS}, demonstrating the paramount importance of system clarification in dialog-based semantic parsing and the database querying process in general.
Another 20\% of this subgroup of dialogues possess questions that cannot be answered without the aid of human inference.
These proportions of \texttt{AMBIGUOUS} and \texttt{INFER\_SQL} identify them the most frequently occurring user dialogue acts outside of \texttt{INFORM\_SQL} and \texttt{THANK\_YOU}, testifying to the diverse complexity of \Ours{}.
}

\paragraph{Semantic complexity}

\begin{figure}[!t]
    \centering
    \includegraphics[width=0.5\textwidth]{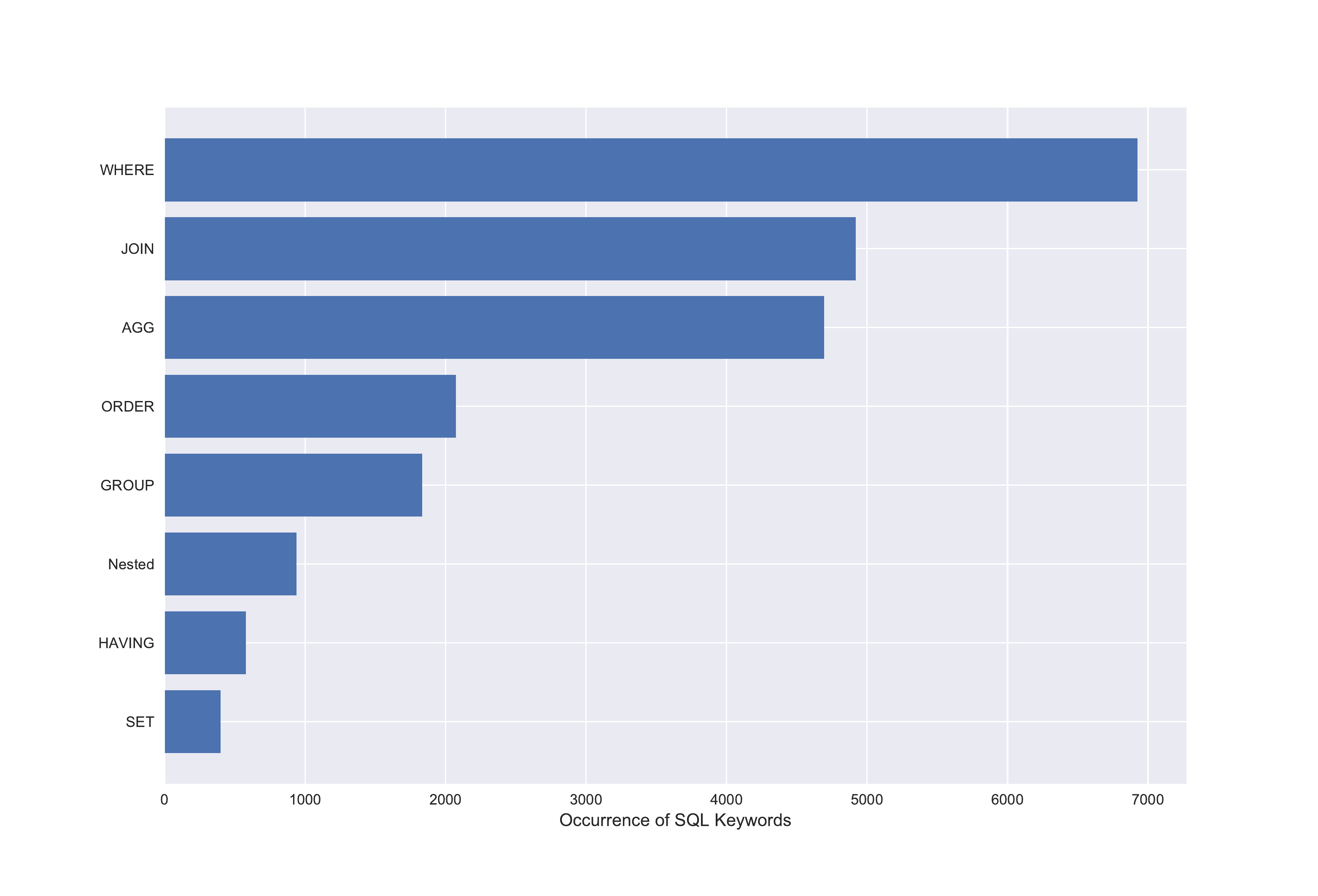}
    \vspace{-2mm}
    \caption{SQL keyword counts.}
\label{fig:sql_count}
\vspace{-2mm}
\end{figure}

\begin{figure}[!t]
    \vspace{-1.5mm}\hspace{-1mm}
    \centering
    % \includegraphics[width=0.48\textwidth]{images/sql_turn_aits.png}
    % \vspace{-2mm}
    % \includegraphics[width=0.48\textwidth]{images/sql_turn.png}
    % \vspace{-2mm}
    \includegraphics[width=0.48\textwidth]{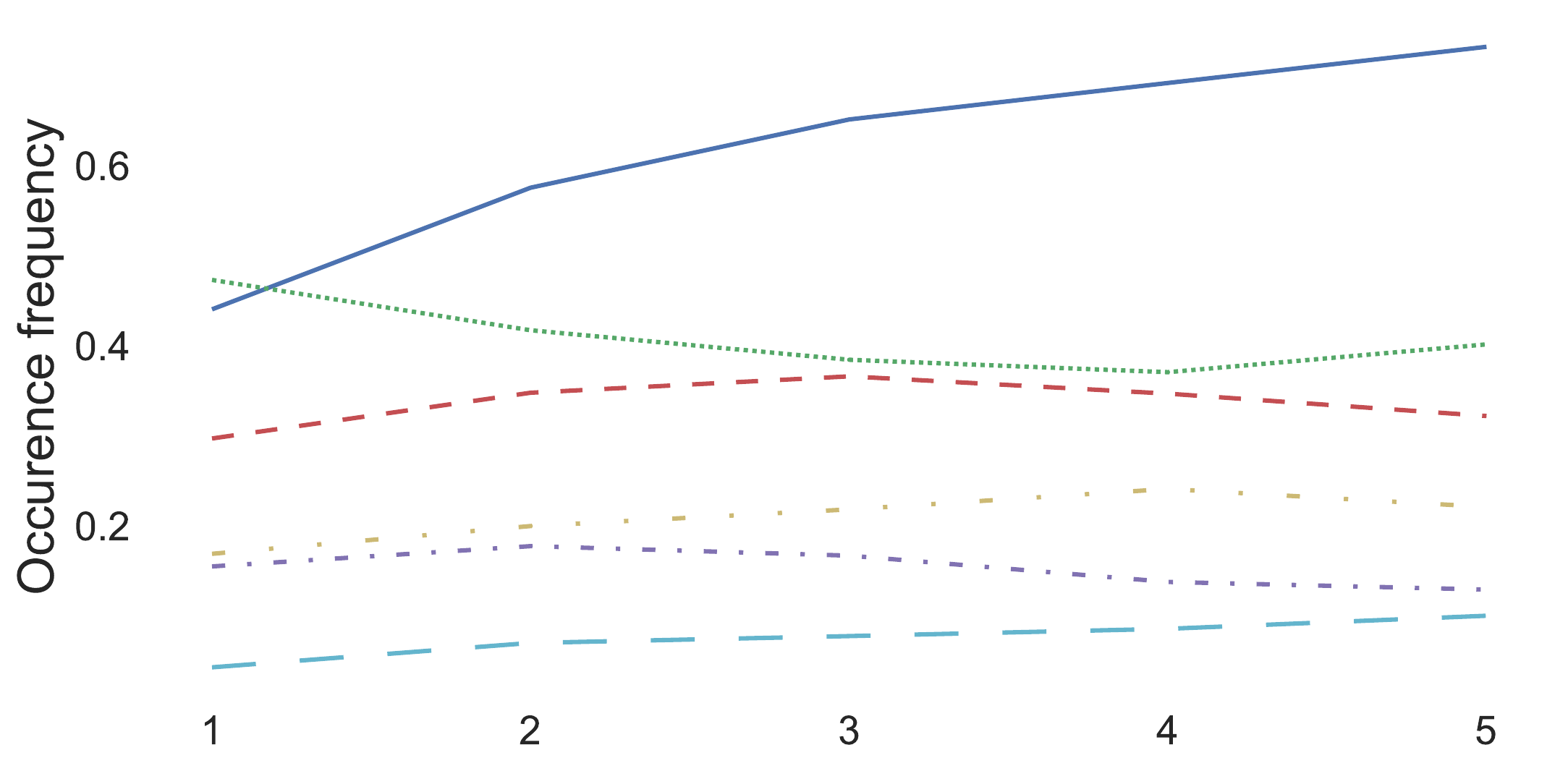}
    \includegraphics[width=0.48\textwidth]{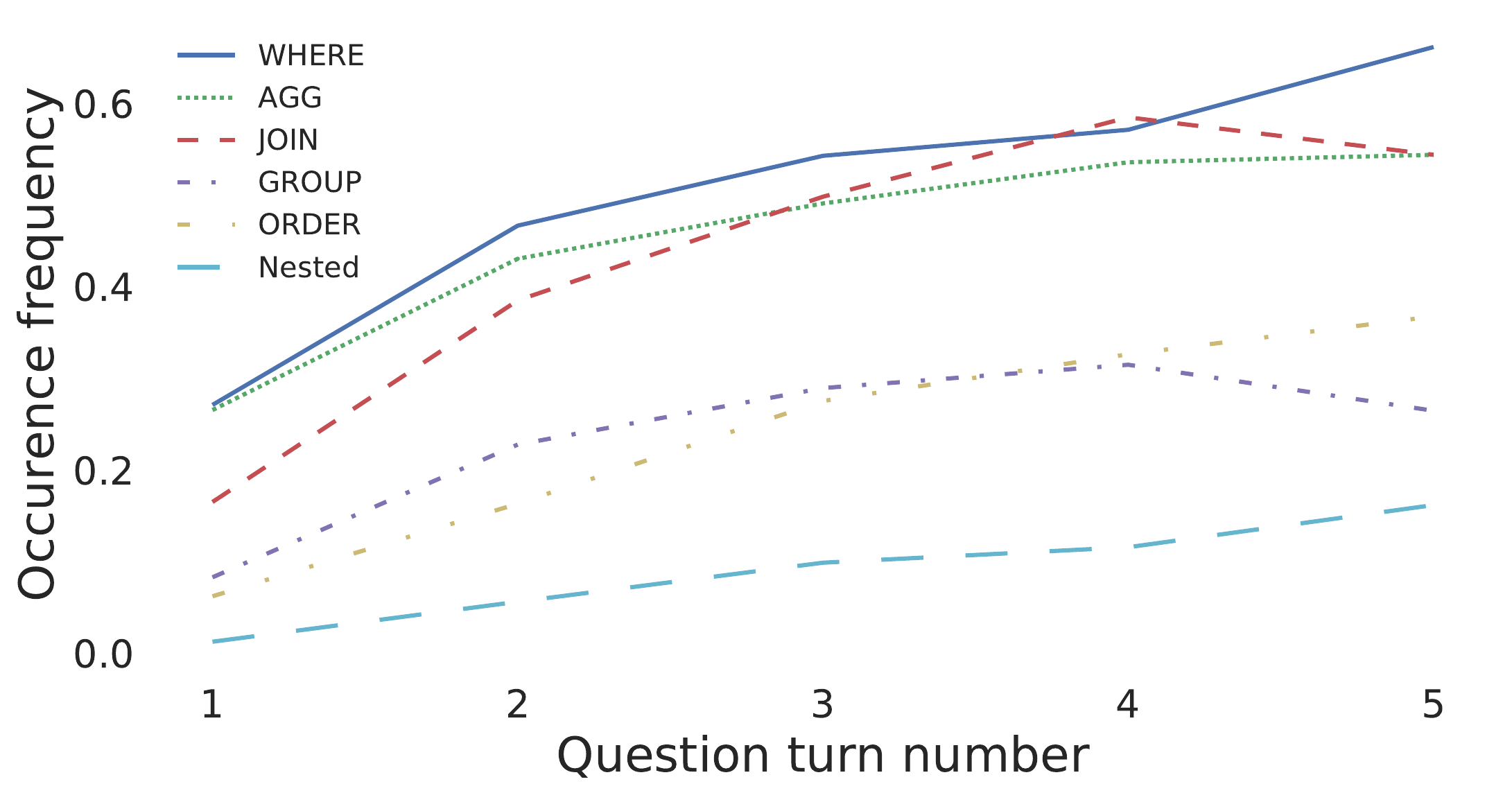}
    \caption{Percentage of question sequences that contain a particular SQL keyword at a specific user utterance turn. The keyword occurrences in \Ours{} (upper) slightly fluctuates as the interaction proceeds while that in SParC (lower) demonstrates a clear increasing trend.
    %\vic{Better to add a title to each subfigure marking the dataset instead of describing them in the caption and the main paragraph.}
    }
    %More complex and specific questions are asked as the number of turns increases in our task, while questions in different turns in ATIS have similar semantic logic.\vic{Nice observation and very interesting characterization. However, need to explain what causes such differences.}
\label{fig:sql_turn}
\vspace{-1mm}
\end{figure}

As shown in Table~\ref{tb:data_stats}, if we consider the column names of the 200 DBs of \Ours{} as slots and their entries as values, the number of slot-value pairs far exceed those defined in other task-oriented dialogues. 
% Furthermore, \Ours{} contains 11,039 questions that can be converted into SQL queries. % stored with this dataset ensure that a wide range of semantic meanings is covered.
Figure \ref{fig:sql_count} shows the total number of occurrences of different SQL keywords in the SQL queries corresponding to these questions.
The SQL queries in \Ours{} cover all common SQL keywords as well as complicated syntactic structure such as nesting (Figure~\ref{fig:sql_turn}). %, demonstrating a diverse semantic coverage.

\hide{
Considering the column names of the 200 databases within this corpus as slots and their entries as values, the number of each possessed by \Ours{} far surpasses that of other dialog datasets, an indication that the corpus retains much more powerful semantic meaning representation ability compared to others of its kind.
This is shown in Table \ref{tb:data_stats}.
Furthermore, the 11,039 questions that can be converted into SQL queries stored with this dataset ensure that a wide range of semantic meanings is covered.
Figure \ref{{fig:sql_count}} shows the total number of occurrences for different SQL keywords in the SQL queries corresponding to these questions.
The SQL queries in our dataset, some of which are nested, cover all common SQL keywords, demonstrating that the semantic coverage of questions is vast and diverse.
}

\paragraph{Semantic changes by turns}

We compute the frequency of occurrences of common SQL keywords in different turns for both \Ours{} and SParC and compare them in Figure~\ref{fig:sql_turn} (upper: \Ours{}, lower: SParC). Here we count the turn \# based on user utterance only. Since \Ours{} and SParC span the same domains, Figure~\ref{fig:sql_turn} reveals a comparison of semantic changes between context-dependent DB questions issued by end users (\Ours{}) and expert users (SParC).
For \Ours{}, the frequencies of all keywords except for \texttt{WHERE} do not change significantly throughout the conversation, and the average frequencies of these keywords are in general lower than those of SParC. In addition, \texttt{WHERE} occurs slightly more frequently in \Ours{} than in SParC. We believe this indicates the exploratory nature of the dialogues we collected, as the users switch their focus more frequently instead of building questions upon previous ones. For example, SQL \texttt{AGG} components occur most frequently in the beginning of dialogues, % aligning with 
as a result of users familiarizing themselves with the amount of data in the DB or other statistical measures. 
In contrast, the frequencies % of occurrence 
of almost all SQL components in SParC increase as the question turn increases. 
This suggests that questions in SParC have stronger inter-dependency, % are more inter-related
as the purpose of this corpus is to study text-to-SQL in context.

\paragraph{Cross domain}
As shown in Table \ref{tab:data_split}, the dialogues in \Ours{} are randomly split into train, development and test sets by DB with a ratio of 7:1:2 (the same split as SParC and Spider).

\begin{table}[h!]
\centering
\small
\begin{tabular}{cccc}
\Xhline{2\arrayrulewidth}
 & Train & Dev & Test \\\hline
\# Dialogs & 2164 & 292 & 551\\
\# Databases & 140 & 20 & 40 \\
\Xhline{2\arrayrulewidth}
\end{tabular}
\caption{Dataset Split Statistics}
\label{tab:data_split}
\end{table}

\hide{
The corpus is randomly split into train, development and test sets. \vic{Data split stats table?}
Each dialogue consists of a user's query goal that is a question in the Spider, multiple user and system utterances, SQL and dialogue acts per interaction, database id, SQL answer of the final query goal.
}

\hide{
% Victoria: probably no space for this study in the main paper?
\paragraph{Qualitative study}
In this section, we provide some examples to conduct qualitative study of \Ours{}.
}
\section{Tasks and Models}
\label{sec:task}

\Ours{} is meant to be used as the first benchmark for building general-purpose DB querying dialogue systems in arbitrary domains. 
Such systems take a user question and determine if it can be answered by SQL (user dialogue act prediction).
If the question can be answered by SQL, the system translates it into the corresponding SQL query (SQL-grounded dialogue state tracking), executes the query, returns and shows the result to the user.
To improve interpretability and trustworthiness of the result, the system describes the predicted SQL query and result tables to the user for their verification (response generation from SQL and result).
Finally, the user checks the results and the system responses and decides if the desired information is obtained or additional questions shall be asked.% clarify, rephrase their questions again.

Some components relevant to the process above are beyond the scope of our work. 
First, our response generation task only includes turns % does not consider generating a response cases 
where the system's dialogue act is \texttt{CONFORM\_SQL}.
In case the system cannot understand the user's question (the system dialogue act is \texttt{CANNOT\_ANSWER}) or considers it as unanswerable (\texttt{CANNOT\_ANSWER}), the system will reply in a standard way to inform the user that it needs clarification or cannot answer that question.
The same applies to questions that require human inference (e.g., the system confirms with the user which types of dorms he or she was talking about by asking $R_3$ instead of immediately translating $Q_3$ in Figure~\ref{fig:task}).
% A simple solution is to ask the user to be more specific or to rephrase the question.
Currently we do not have a task setup to evaluate the quality of system clarifications.
Second, some user questions cannot be directly answered by SQL but are possible to be answered with other type of logical reasoning  (e.g., $Q_3$ in Figure~\ref{fig:dialogue_example_1}).
% We excluded them from the SQL-grounded dialogue state tracking task 
We exclude these questions from our task design and leave them for future research.

\subsection{SQL-Grounded dialogue State Tracking}
\label{sec:dstc}

In \Ours{}, user dialogue states are grounded in SQL queries. % instead of slot-value pairs in predefined task-specific ontology.
Dialogue state tracking (DST) in this case is to predict the correct SQL query for each user utterance % question 
with \texttt{INFORM\_SQL} label given % the entire history of user questions 
the interaction context and the DB schema.
In our setup, the system does not have access to gold SQL queries from previous turns, which is different from the traditional DST settings in dialogue management where the history of ground-truth dialogue states is given. % as an input to predict the SQL query for the current user question.
% Given the user questions with \texttt{INFORM\_SQL} label in a dialogue and the corresponding DB schema, the task is to generate SQL queries for each question.
Comparing to other context-dependent text-to-SQL tasks such as SParC %\cite{Yu2019} 
and ATIS, %\cite{Suhr:18context} 
the DST task in \Ours{} also include the ambiguous questions if the user affirms the system clarification of them (e.g., $Q_4$ in Figure~\ref{fig:task}).
%\vic{Are there any ambiguous questions not confirmed by the user? If so, what is the ratio of such questions? Will they be included in the coversation history or not?}
In this case, the system clarification is also given as part of the interaction context  % also considers the system's response 
to predict the SQL query corresponding to the question.\footnote{If a dialogue contains multiple ambiguous questions, the system clarification to all ambiguous questions will be given as input.}\footnote{The ambiguous questions not confirmed by the user and their system responses are given as part of the conversation history but we do not require a system to predict SQL queries for them.}
%\vic{Okay, this one confuses me significantly. I thought the task is simply to give model access to all previous interaction history hence system response should certainly be there. What is special about this case? To describe it better, you probably need a full example, i.e. explicitly list here what information the model can have access to using figure 1}
For instance, to generate $S_4$ in Figure~\ref{fig:task}, the input consists of all previous questions ($Q_1, Q_2, Q_3$), the current user question ($Q_4$), the DB schema, and the system response $R_3$.
%Finally, for this task, we have 2159/293/548 dialogues with 7343/1007/1904 SQL-question pairs for train/development/test sets \footnote{7 dialogues are excluded because none of the questions cannot be answered by SQL queries}.

We benchmark the performance of two % state-of-the-art 
strong context-dependent neural text-to-SQL models on this the task, which are the baseline models reported on SParC by~\newcite{Yu2019}.
% They are extended for the SParC task \cite{Yu2019}.% whose setting is the same as our task.

\paragraph{Context-dependent Seq2Seq (CD-Seq2Seq)} 
The model is originally introduced by~\cite{Suhr:18context} for the ATIS task.
It incorporates interaction history and is able to copy segments of previous generated SQL queries.
\citet{Yu2019} extends it to encode DB schema information such that it works for the cross-domain setting in SParC.
We apply the model to our task without any changes.

\paragraph{SyntaxSQL-con} 
SyntaxSQLNet is a SQL-specific syntax-tree based model introduced for Spider
% , the complex, cross-domain text-to-SQL task
~\cite{Yu&al.18.emnlp.syntax}.
\citet{Yu2019} extends it to take previous questions as input when predicting SQL for the current question.
We apply the model to our task without any changes.

\subsection{Response Generation from SQL and Query Results}
\label{sec:datg}

This task requires generating a natural language description of the SQL query and the result for each system response labeled as \texttt{CONFORM\_SQL}.
% The data in the task is SQL queries and execution results in \Ours{} if their system responses is labelled as \texttt{CONFORM\_SQL}.
It considers a SQL query, the execution result, and the DB schema.
% Comparing to other data-to-text generation tasks, 
Preserving logical consistency between SQL and NL response is crucial in this task, in addition to naturalness and syntactical correctness.
Unlike other SQL-to-text generation tasks \cite{xu2018sql2text}, our task maps the SQL query to a statement and summarizes the result in that statement (instead of just mapping it back to the user question).

% To benchmark the challenges of this task, 
We experiment with three baseline methods for this task.
\paragraph{Template-based}
% We first create a list of SQL-response one-to-one mapping between SQL and responses without any specific values, column and table names.
% We first curate a large set of SQL-response pairs from the collected data. 
Given the SQL and NL response pairs in the training set, we masked variable values in both the SQL and NL response % in this set 
to form parallel SQL-response templates.
Given a new SQL query, we employ rule-based approach to select the closest SQL-response template pair from the set.
After that, we fill in the selected response template with the columns, tables, and values of the SQL query and the result to generate the final response (see more in Appendix).

\paragraph{Seq2Seq}
We experiment with a vanilla Seq2Seq model~\cite{sutskever2014sequence} with attention~\cite{bahdanau2015neural}, a standard baseline for text generation tasks. 

\paragraph{Pointer-generator}
Oftentimes the column or table names in the NL response are copied from the input SQL query. To capture this phenomenon, we experiment with a pointer-generator network~\cite{See2017}, which addresses the problem of out-of-vocabulary word generation in summarization and other text generation tasks. We use a modified version of the % maximum likelihood-trained 
implementation from \newcite{chen2018fast}.

\subsection{User dialogue Act Prediction}
\label{sec:ddtg}

\begin{table*}[ht!]
\centering
\scalebox{0.8}{
\begin{tabular}{p{20mm}p{158mm}}
\Xhline{2\arrayrulewidth}
Groups & Dialog acts   \\ \hline
%the subset, superset or disjoint set of
DB user &  \texttt{inform\_sql}, \texttt{ infer\_sql}, \texttt{ ambiguous}, \texttt{ affirm}, \texttt{ negate}, \texttt{ not\_related}, \texttt{ cannot\_understand}, \texttt{ cannot\_answer}, \texttt{ greeting}, \texttt{ goodbye}, \texttt{ thank\_you} \\ \hline
DB expert &  \texttt{conform\_sql}, \texttt{clarify}, \texttt{reject}, \texttt{request\_more}, \texttt{greeting}, \texttt{sorry}, \texttt{welcome}, \texttt{goodbye} \\
\Xhline{2\arrayrulewidth}
\end{tabular}}
\caption{Dialog acts in \Ours{}. See \S~\ref{sec:dialog-acts} for the comprehensive definition of each dialogue act.} 
\label{tb:acts}
\end{table*}

For a real-world DB querying dialogue system, it has to decide if the user question can be mapped to a SQL query or if special actions are needed.
We define a series of dialogue acts for the DB user and the SQL expert (Table~\ref{tb:acts}).\footnote{\S\ref{sec:dialog-acts} defines the complete set of dialogue action types.}
% Given a user question, the task maps it to either one or more user dialogue acts defined in.
For example, if the user question can be answered by a SQL query, the dialogue act of the question is \texttt{INFORM\_SQL}.
% \texttt{INFER\_SQL} means the user question must be answered by a SQL query plus human inference.
Since the system DATs % s of the DB querying dialogue system is 
are defined in response to the user DATs, our task does not include system dialogue acts prediction.
% For example, in Figure \ref{fig:task}, if the user's dialogue act is \texttt{INFORM\_SQL}, the system's dialogue act will be  \texttt{CONFIRM\_SQL}, and it parses the question into a SQL query.
% If not, the system needs to decide to which user's dialogue act belongs to, and then choose an appropriate action to generate a response.

We experiment with two baseline models for this task.

\paragraph{Majority} 
The dialogue acts of all the user questions are predicted to be the majority dialogue act \texttt{INFORM\_SQL}.

\paragraph{TBCNN-pair} 
We employ TBCNN-pair~\cite{Mou2016}, a tree-based % convolutional neural 
CNN model with heuristics for predicting entailment and contradiction between sentences.
We change the two sentence inputs for the model to a user utterance and the DB schema, and follow the same method in SQLNet~\cite{Xu2017} to encode each column name.
\section{Results and Discussion}
\label{sec:result}

\paragraph{SQL-grounded dialog state tracking}

\begin{table}[h!]
\centering
\resizebox{\columnwidth}{!}{
\begin{tabular}{ccccc}
\Xhline{2\arrayrulewidth}
Model & \multicolumn{2}{c}{Question Match} & \multicolumn{2}{c}{Interaction Match} \\
              & Dev & Test &  Dev & Test \\ \hline
\seqcon{}          & 13.8 & 13.9 & 2.1 & 2.6 \\
\syncon{} & 15.1 & 14.1 & 2.7 & 2.2 \\
\Xhline{2\arrayrulewidth}
\end{tabular}
}
\caption{Performance of various methods over all questions (\textit{question match}) and all interactions (\textit{interaction match}).} 
\label{tb:res_sql}
\end{table}

We use the % exact set matching accuracy 
same evaluation metrics used by the SParC dataset~\cite{Yu2019} to evaluate the model's performance on all questions and interactions (dialogs).
The performances of CD-Seq2Seq and SyntaxSQL-con are reported in Table~\ref{tb:res_sql}.
The two models achieve less than 16\% % on the accuracy of SQL matching on all the questions and less than 3\% on all the dialogs.
question-level accuracy and less than 3\% on interaction-level accuracy.
Since the two models have been benchmarked on both CoSQL and SParC, we cross-compare their performance on these two datasets.
Both models perform significantly worse on  \Ours{} DST than on SParC.
This indicates that \Ours{} DST is more difficult than SParC.
% the difficulty of \Ours{} is harder than SParC.
The possible reasons is that the questions in \Ours{} are generated by a more diverse pool of users (crowd workers instead of SQL experts), the task includes ambiguous questions and the context contains more complex intent switches.
% The reason might be because of the richer diversity of users and dialogs (such as mixing with intent switches and inter-related questions in a single dialog, and ambiguous questions).

\paragraph{Response generation}

\begin{table}[h!]
\centering
\resizebox{\columnwidth}{!}{
\begin{tabular}{ccccc}
\Xhline{2\arrayrulewidth}
Model & \multicolumn{2}{c}{BLEU} & LCR (\%) & Grammar\\
              & Dev & Test &  Test & Test\\ \hline
Template      & 9.5 & 9.3 & 41.0 & 4.0 \\
Seq2Seq       & 15.3 & 14.1 & 27.0 & 3.5 \\
Pointer-generator & 16.4 & 15.1 & 35.0 & 3.6 \\
\Xhline{2\arrayrulewidth}
\end{tabular}
}
\caption{BLEU scores on the development and test sets, and human evaluations of logic correctness rate (LCR) and grammar check on the 100 examples randomly sampled from the test set.} 
\label{tb:res_response}
\end{table}

Table~\ref{tb:res_response} shows the results of three different baselines on three metrics: BLEU score~\cite{Papineni2002}, logic correctness rate (LCR), and grammar.
To compute LCR and grammar score, we randomly sampled 100 descriptions generated by each model. 
Three students % with English professions 
proficient in English participated in the evaluation,
They were asked to choose a score 0 or 1 for LCR, and 1 to 5 for grammar check (the larger, the better).
For LCR, the final score was decided by majority vote.
We computed the average grammar score.

Interestingly, the human evaluation and BLEU scores % lead to the opposite results on template and neural based models.
do not completely agree.
While the template-based method is brittle and requires manual effort, % tedious and unscalable,
it performs significantly better than the two end-to-end neural models in the human evaluation.
Because the SQL-question templates provide 
% the basic skeleton to generate the 
natural and grammatical sketch of the output, it serves as an advantage in our % task for the 
human evaluation.
% Considering the complexity of SQL queries including nested queries in \Ours{}, it is impossible to have a template covering all SQL queries especially when testing on different DBs.
However, this approach is limited by the small coverage of the training templates and its LCR is only around 40\%.
On the other hand, the neural models achieve better BLEU scores than the template-based approach.
A possible reason for this is that they tend to generate words frequently associated with certain SQL queries.
However, the neural models struggle to preserve the SQL query logic in the output.
Unsurprisingly, pointer-generator performs better than basic Seq2Seq in terms of both BLEU and human evaluation.
The low performances of all methods on LCR show that the task % built on \Ours{} 
is indeed very challenging.

\paragraph{User dialog act prediction}

\begin{table}[h!]
\centering
\small
\begin{tabular}{ccc}
\Xhline{2\arrayrulewidth}
Model & Dev & Test \\\hline
Majority  & 63.3 & 62.8  \\
TBCNN-pair  & 84.2 & 83.9 \\
\Xhline{2\arrayrulewidth}
\end{tabular}
\caption{Accuracy of user dialog act prediction on the development and test sets.}
\label{tb:res_intent}
\end{table}

Table~\ref{tb:res_intent} shows the accuracy of the two baselines on predicting user dialog acts.
The result of Majority indicates that about 40\% of user questions cannot be directly converted into SQL queries.
This confirms the necessity of considering a larger set of dialogue actions for building a practical NLIDB system.
Even though TBCNN can predict around 85\% of user intents correctly, most of the correct predictions are for simple classes such as \texttt{INFORM\_SQL}, \texttt{THANK\_YOU}, and \texttt{GOODBYE} etc.
The F-scores for more interesting and important dialog acts such as \texttt{INFER\_SQL} and \texttt{AMBIGUOUS} are around 10\%.
This indicates that improving the accuracy on user DAT prediction is still important.
% also important to build a user-friendly DB query system.
\section{Conclusion and Future Work}
\label{sec:conclusion}
In this paper, % we introduce \Ours{}, a large scale dataset for building a cross-domain, general-purpose DB querying system.
we introduce \Ours{}, the first large-scale cross-domain conversational text-to-SQL corpus collected under a Wizard-of-Oz setup.
% Unlike existing text-to-SQL datasets, \Ours{} was collected through dialogs between a DB user with diverse background and a SQL expert querying a large amount of different DBs.
Its language and discourse diversity and cross-domain setting raise exciting open problems for future research. 
Especially, the baseline model performances on the three challenge tasks suggest plenty space for improvement. 
The data and challenge leaderboard will be publicly available at \url{https://yale-lily.github.io/cosql}.

\paragraph{Future Work} As discussed in Section~\ref{sec:task}, some examples in \Ours{} include ambiguous and unanswerable user questions and we do not study how a system can % generate effective responses to 
effectively clarify those questions or guide the user to ask questions that are answerable.
Also, some user questions cannot be answered with SQL but by other forms of logical reasoning the correct answer can be derived.
We urge the community to investigate these problems in future work in order to build practical, robust and reliable conversational natural language interfaces to databases.

\section{Acknowledgement}
%To thank IBM/Shappire project and many others
This work was supported in part by IBM under the Sapphire Project at the University of Michigan.
% We thank the anonymous reviewers for their thoughtful detailed comments.

\bibliography{emnlp-ijcnlp-2019}
\bibliographystyle{acl_natbib}

\appendix
\section{Appendices}
\label{sec:appendix}

\begin{figure*}[!t]
    \centering
    \includegraphics[width=0.8\textwidth]{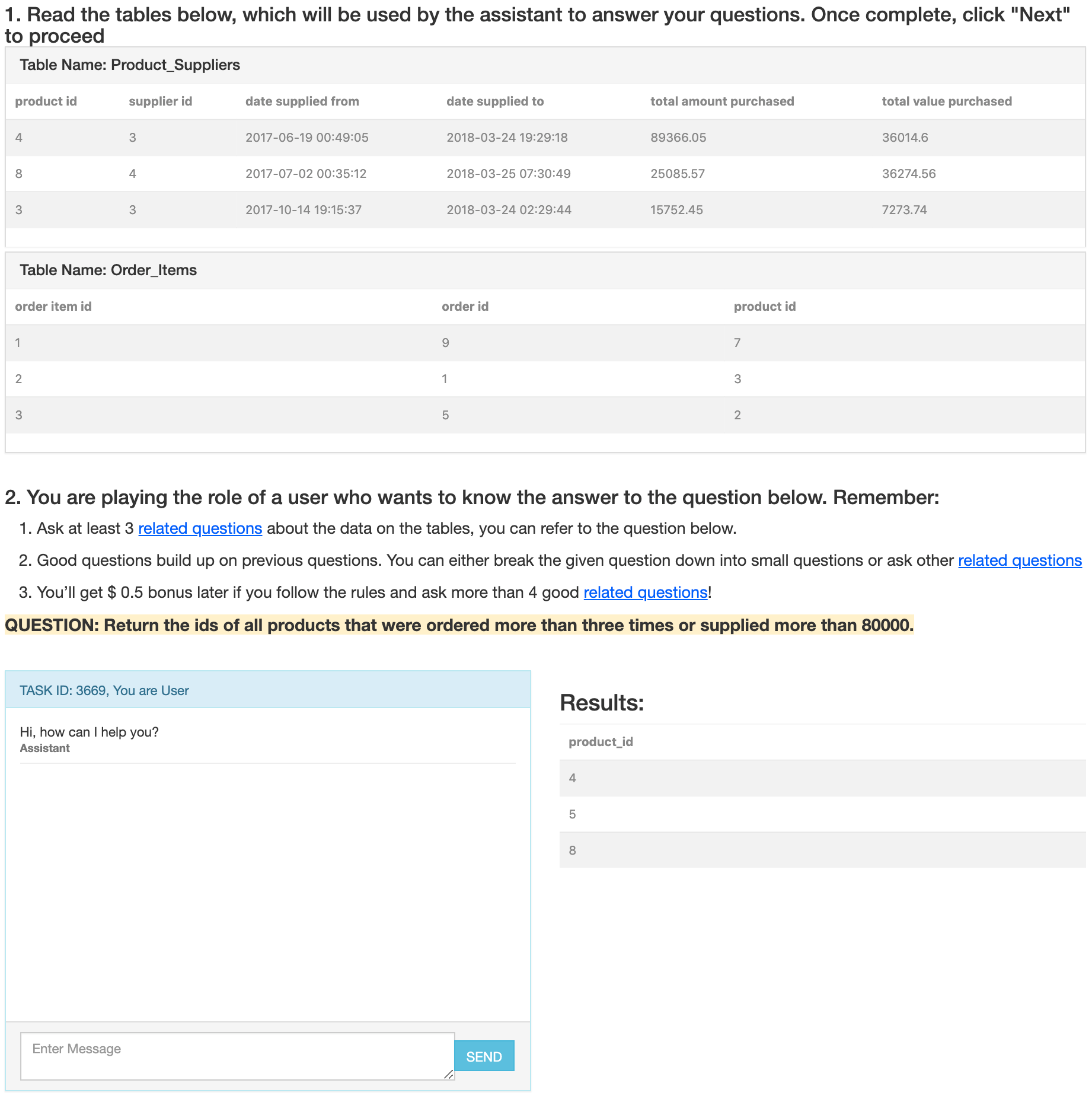}
    
    \caption{DB User Interface}
\label{fig:db_user_interface}

\end{figure*}

\begin{figure*}[!t]
    \centering
    \includegraphics[width=0.8\textwidth]{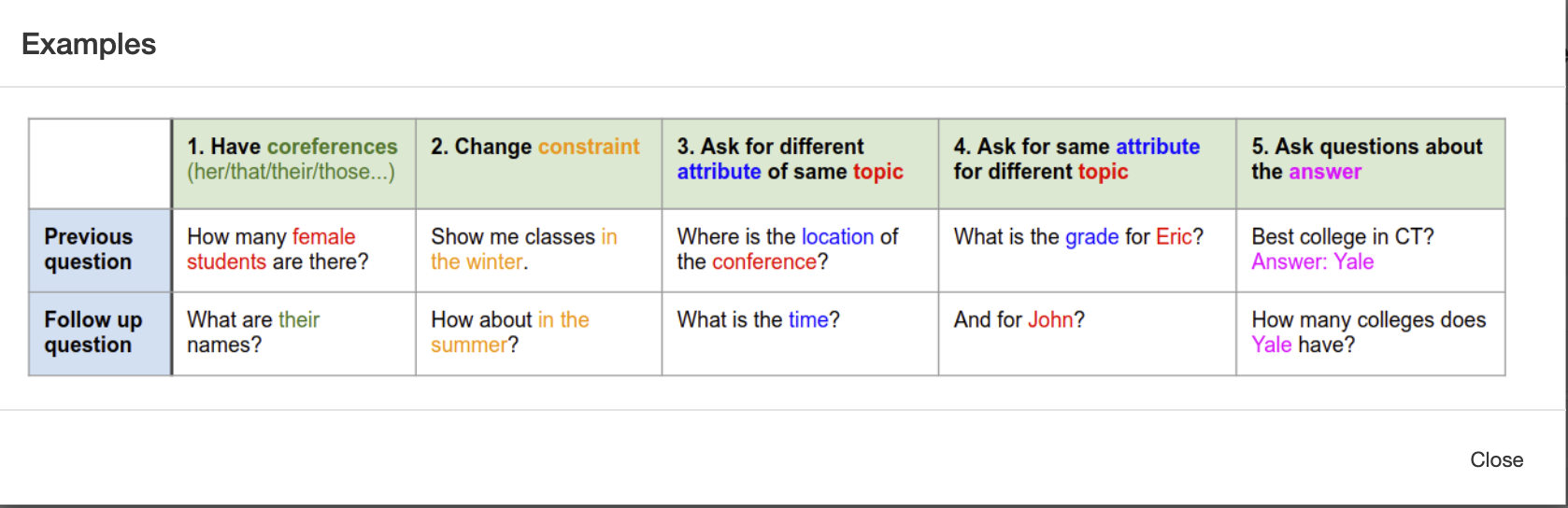}
    
    \caption{DB User Related Questions: a pop-up window when the user clicks highlighted "related questions" in the above interface.}
\label{fig:db_user_related_questions}

\end{figure*}

\begin{figure*}[!t]
    \centering
    \includegraphics[width=0.8\textwidth]{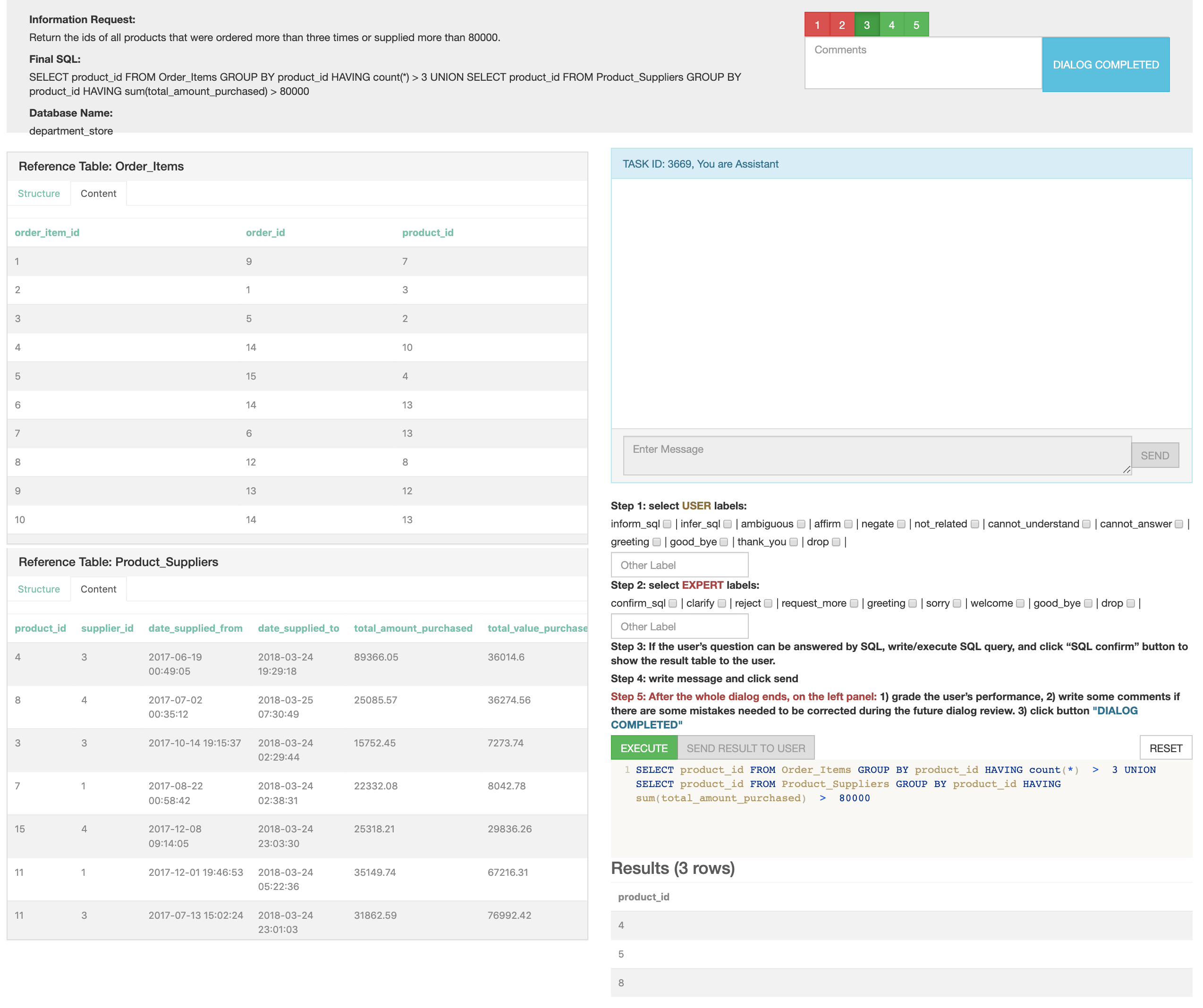}
    \caption{SQL Expert Interface}
\label{fig:sql_expert_interface}
\end{figure*}

\begin{figure*}[!t]
    \centering
    \includegraphics[width=0.8\textwidth]{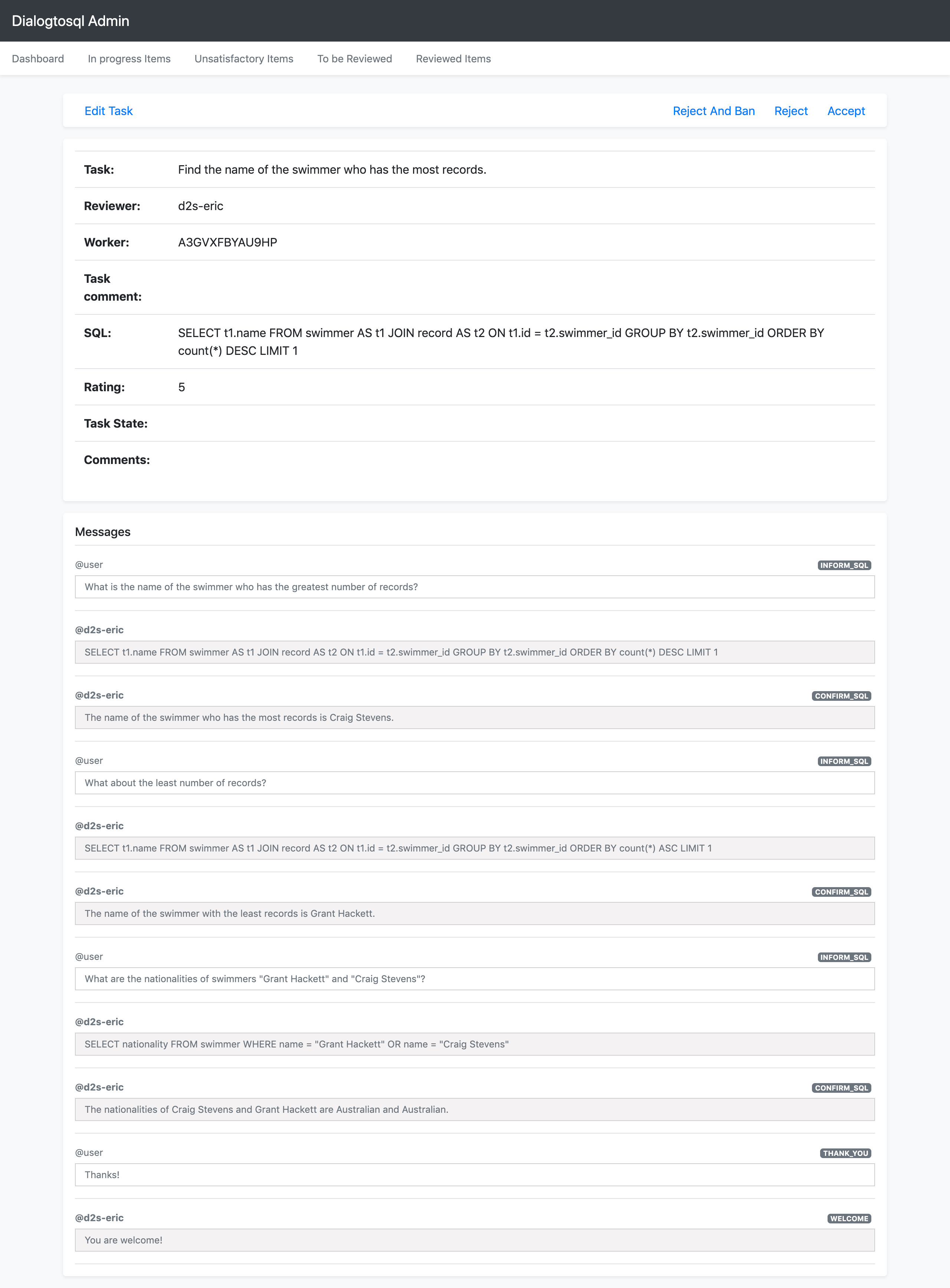}
    \caption{Dialogue Review Interface}
\label{fig:review_interface}
\end{figure*}

\begin{figure*}[!t]
    \centering
    \includegraphics[width=0.48\textwidth]{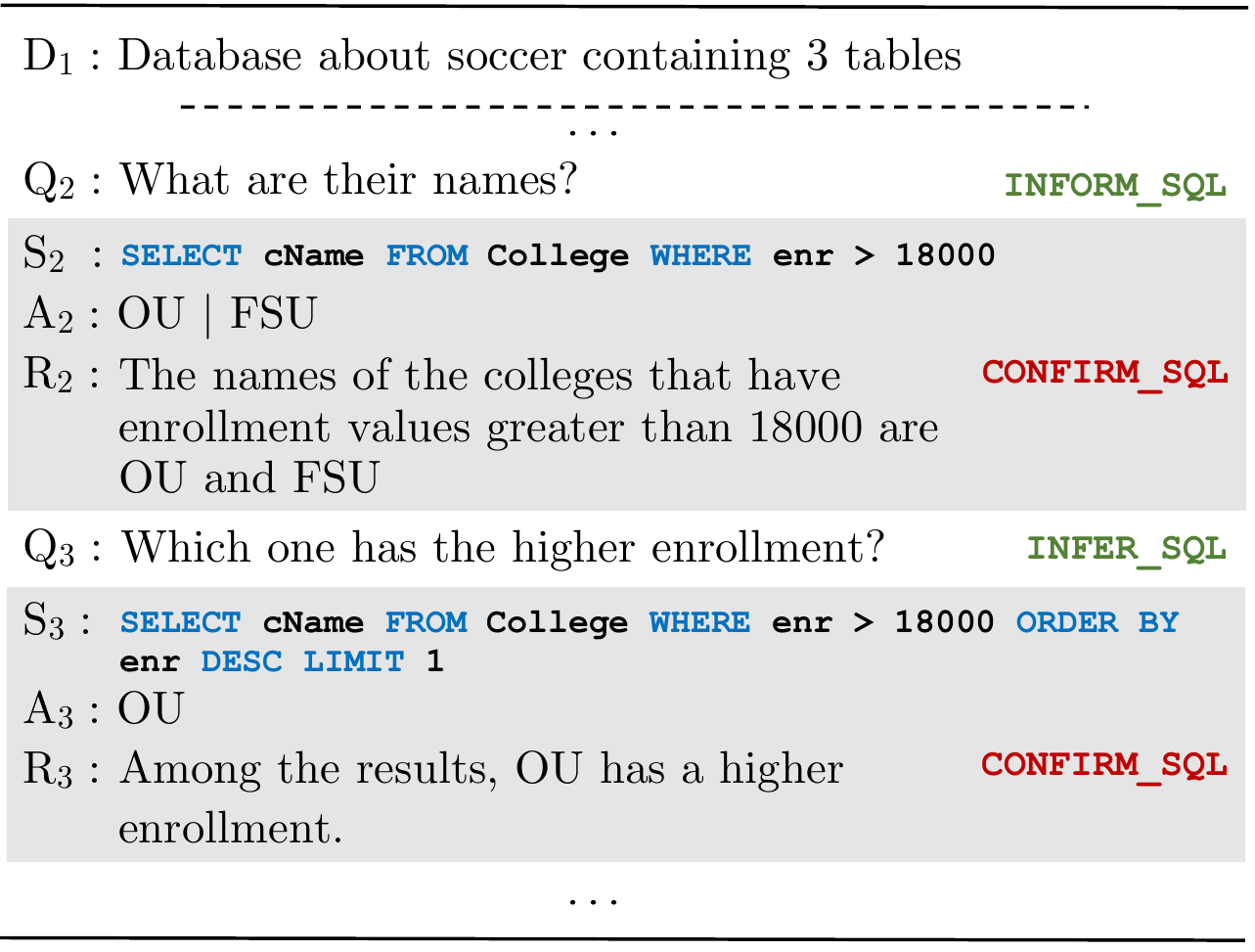}
    \caption{Part of a dialogue example with \textsc{INFER\_SQL} user dialog label}
\label{fig:dialogue_example_1}
\end{figure*}

\begin{figure*}[!t]
    \centering
    \includegraphics[width=0.48\textwidth]{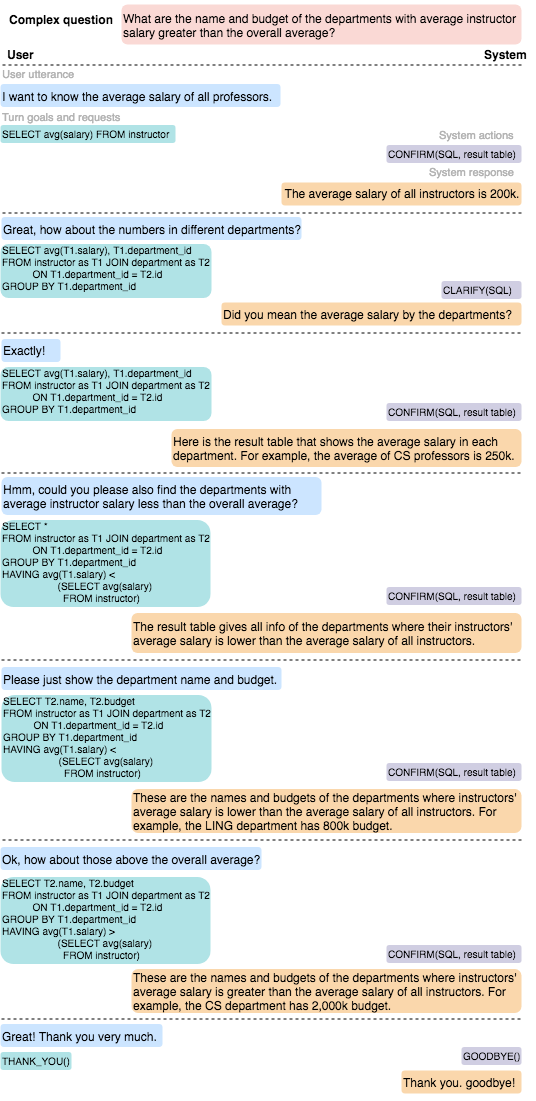}
    \caption{A complete dialogue example}
\label{fig:dialogue_example_2}

\end{figure*}

This section provides description of dialog actions in A.1, more details on baseline modifications and hyperparameters in A.2, system response guides in A.3, additional dialog examples in CoSQL dataset in Figure \ref{fig:dialogue_example_1} and \ref{fig:dialogue_example_2}, and the DB user (AMT turkers) and the SQL expert (college computer science students) annotation interfaces in Figure \ref{fig:db_user_interface}, \ref{fig:db_user_related_questions}, \ref{fig:sql_expert_interface}, and \ref{fig:review_interface}.

\subsection{Description of Dialog Acts}
\label{sec:dialog-acts}
For the DB user, we define the following dialog acts:

\paragraph{\texttt{INFORM\_SQL}}
The user informs his/her request if the user’s question can be answered by SQL. The system needs to write SQL.
\paragraph{\texttt{INFER\_SQL}}
If the user’s question must be answered by SQL+human inference. For example, users’ questions are “are they..?” (yes/no question) or “the 3rd oldest...”. SQL cannot directly (or unnecessarily complicated) return the answer, but we can infer the answer based on the SQL results.
\paragraph{\texttt{AMBIGUOUS}}
The user’s question is ambiguous, the system needs to double check the user's intent (e.g. what/did you mean by...?) or ask for which columns to return.
\paragraph{\texttt{AFFIRM}}
Affirm something said by the system (user says yes/agree).
\paragraph{\texttt{NEGATE}}: 
Negate something said by the system (user says no/deny).
\paragraph{\texttt{NOT\_RELATED}}
The user’s question is not related to the database, the system reminds the user.
\paragraph{\texttt{CANNOT\_UNDERSTAND}}
The user’s question cannot be understood by the system, the system asks the user to rephrase or paraphrase question.
\paragraph{\texttt{CANNOT\_ANSWER}}
The user’s question cannot be easily answered by SQL, the
system tells the user its limitation.
\paragraph{\texttt{GREETING}}
Greet the system.
\paragraph{\texttt{GOOD\_BYE}}
Say goodbye to the system.
\paragraph{\texttt{THANK\_YOU}}
Thank the system.
\\
\\
For the system, we define the following dialog acts:

\paragraph{\texttt{CONFIRM\_SQL}}
The system creates a natural language response that describes SQL
and result table, and asks the user to confirm if the system understood his/her intention.
\paragraph{\texttt{CLARIFY}}
Ask the user to double check and clarify his/her intention when the user’s question is ambiguous.
\paragraph{\texttt{REJECT}}
Tell the user you did not understand/cannot answer his/her question, or the user question is not related.
\paragraph{\texttt{REQUEST\_MORE}}
Ask the user if he/she would like to ask for more info.
\paragraph{\texttt{GREETING}}
Greet the user.
\paragraph{\texttt{SORRY}}
Apologize to the user.
\paragraph{\texttt{WELCOME}}
Tell the user he/she is welcome.
\paragraph{\texttt{GOOD\_BYE}}
Say goodbye to the user.

\subsection{Modifications and Hyperparameters for Baselines}

\paragraph{CD-Seq2Seq}
We apply the model with the same settings used in SParC without any changes.

\paragraph{SyntaxSQL-con}
We apply the model with the same settings used in SParC without any changes.

\paragraph{Template-based}
We first create a list of SQL query patterns without values, column and table names that cover the most cases in the train set of CoSQL. 
And then we manually changed the patterns and their corresponding responses to make sure that table, column, and value slots in the responses have one-to-one map to the slots in the SQL query.
Once we have the SQL-response mapping list, during the prediction, new SQL statements are compared with every templates to find the best template to use. A score will be computed to represent the similarity between the SQL and each template.
The score is computed based on the number of each SQL key components existing in the SQL and each template. Components of the same types are grouped together to allow more flexible matching, like count, max, min are grouped to aggregate.
A concrete example of templates is shown: 
{\small\tt SELECT column0 FROM table0 WHERE column1 comparison0 value0}. {\small\tt column{0,1}} and {\small\tt table0} represent column name and table name respectively. {\small\tt comparison0} represents one of the comparison operator including {\small\tt >=, <=, <,>,=,!=, and like}. {\small\tt value0} represents a value the user uses to constrain the query result.

\paragraph{Seq2Seq}
We train a word2vec embedding model on the concatenation of the SQL query and response output of the training data for the embedding layer of our Seq2Seq model. We use an embedding dimension of 128, hidden dimension of 256, a single-layer bi-directional LSTM encoder and uni-directional LSTM decoder with attention. We use a batch size of 32, clip the norm of the gradient at 2.0, and do early stopping on the validation loss with a patience of 5. We perform decoding with greedy search. 
\paragraph{Pointer-generator}
We follow the same settings as in the Seq2Seq case with the addition of the copy mechanism during training and testing. 

\paragraph{TBCNN-pair}
The model is modified mainly on the sentence embedding part and classifier part. The input of the modified model is a user utterance and the related column names. Therefore, we replace one of the two sentence embedding modules with a database column name encoding module, which generates representations of the column names related to the sentence. The classifier is modified by adding a label(user dialogue act) number predicting module, which predicts the number of the labels(user dialogue acts) of the user utterance. The label number prediction module is similar to the column number prediction module in SQLNet. 

\subsection{System Response Guide}
System response should be standard and professional. We follow the rules below to write responses for different system dialog action type:

\paragraph{\texttt{CLARIFY}}
"Did you mean...?", "What did you mean by...?", or anything similar.
\paragraph{\texttt{REJECT}}
"Sorry, I don't have the answer to your question..." or anything similar.
\paragraph{\texttt{REQUEST\_MORE}}
"Do you want anything else?" or anything similar.
\paragraph{\texttt{CONFORM\_SQL}}
We convert SQL written by us back to natural language. (We should use the column names and values in the SQL). Our response has to describe all information in the SQL independently instead of referring to any previous context or subject.
\begin{enumerate}
	\item If the returned result can be combined with the SQL description, combine them together to generate the response. For example: 
	
	Given SQL: \\
	{\small \textsc{SELECT avg(salary) FROM instructor}} \\
	Result Returned: 200k \\
	Your Response:``The average salary of all instructors is 200k."
    \item If the returned result is too large and cannot be combined with the SQL description, describe them separately. For example: 
    
    Given SQL: \\
    {\small \textsc{SELECT avg(T1.salary),T1.department\_id FROM instructor as T1 JOIN department as T2 ON T1.department\_id = T2.id GROUP BY T1.department\_id}} \\
    Result Returned: a long table \\
    Your Response:``Here is the result table that shows the average salary in each department. For example, the average of CS professors is 250k."
\end{enumerate}

\end{document}